# Protein Structure Prediction: From Evolutionary Constraints to Generative Modeling

Wengan He[a,b,*,1], Yongsheng Luo[a,1], Lihong Jiang[b], Wenhui Xu[b] and Yu Li[a,*]

[a]Zhuhai College of Science and Technology, Zhuhai, China
[b]Zhuhai UNO Technology Co., Ltd., Zhuhai, China

ABSTRACT

Accurate protein structure prediction is fundamental to structural biology because protein structure underlies molecular function and provides a basis for mechanistic interpretation. Recent advances in deep learning have transformed the field from multiple sequence alignment (MSA)-driven monomer folding into broader frameworks capable of modeling protein complexes and increasingly heterogeneous molecular systems. Existing reviews have summarized this progress from the perspectives of representative models, application domains, and protein design. Building on these efforts, this review focuses on the methodological evolution of the field itself. It examines recent developments through three closely related dimensions: representations and data, architectures and learning strategies, and confidence and evaluation. Within this perspective, the field is organized into four methodological phases and three cross-cutting transitions: from explicit evolutionary coupling features and early contact prediction to learned sequence representations in AlphaFold2, RoseTTAFold, and ESMFold; from protein-only monomer folding to increasingly integrated modeling of heterogeneous molecular systems in AlphaFold-Multimer, RoseTTAFoldNA, and AlphaFold3; and, more recently, from prediction-oriented structure inference to design-oriented generative modeling in RFdiffusion and related frameworks. This framework provides a clearer understanding of how methodological shifts have shaped the capabilities, limitations, and practical roles of recent models.

## 1. Introduction

Protein structure is central to structural biology because it constrains molecular function and enables mechanistic interpretation [6, 18]. The Protein Data Bank provides the principal global archive of experimentally determined macromolecular structures [17]. Traditional computational approaches, including homology modeling, fragment assembly, and physics-based refinement, achieved important progress but remained constrained by template availability, conformational sampling complexity, and the accuracy of energy functions [95, 70, 18].

The emergence of deep learning fundamentally redefined the methodological foundations of structure prediction [66, 59]. Within a short period, the field progressed from co-evolution-assisted contact inference [82] to end-to-end coordinate prediction [7, 47] with near-experimental accuracy for many monomeric proteins [40]. Subsequent developments expanded modeling capabilities to protein–protein complexes [23], protein–nucleic acid assemblies [8], and ligand-inclusive heterogeneous molecular systems [42, 2, 50, 75, 85]. More recently, diffusion- and flow-based generative frameworks [83, 27] extended structural modeling toward controllable protein design.

Existing reviews have summarized this progress from perspectives such as representative model families, application scope, and protein design [59, 63]. The present review instead emphasizes the methodological evolution of protein structure prediction itself. Specifically, it examines recent developments through three closely related dimensions: (1) representations and data, (2) architectures and learning strategies, and (3) confidence and evaluation. As summarized in Figure 1, the field is organized here into four methodological phases spanning explicit evolutionary modeling, end-to-end deep folding, relatively unified complex structure modeling, and generative structure modeling and design. Figure 1 further highlights three associated transitions: from explicit evolutionary coupling features and early contact prediction [82] to learned sequence representations in models such as AlphaFold2 [40], RoseTTAFold [7], and ESMFold [47]; from monomer-focused modeling in the first two phases to increasingly integrated treatment of heterogeneous molecular systems in AlphaFold-Multimer [23], RoseTTAFoldNA [8], RoseTTAFold All-Atom [42], and AlphaFold3 [2]; and, more recently, from prediction-oriented structure inference to design-oriented generative modeling in RFdiffusion [83] and related frameworks [27]. Rather than replacing predictive approaches, generative protein design is treated here as a complementary paradigm that extends structural modeling toward controllable creation and exploration of novel sequence–structure solutions. This phase-and-transition perspective provides a clearer understanding of how methodological shifts have shaped the capabilities, limitations, and practical roles of recent models.

Phases in this review are assigned primarily according to modeling objective and system scope rather than architecture alone. Thus, AlphaFold3 is placed in Phase III because its principal task is structure inference for specified heterogeneous biomolecular assemblies, even though its structure module uses diffusion-based denoising [2].

*Corresponding author.
wenganhe@icloud.com (W. He); jluzhliyu@zcst.edu.cn (Y. Li)
ORCID(s):
[1]These authors contributed equally to this work.

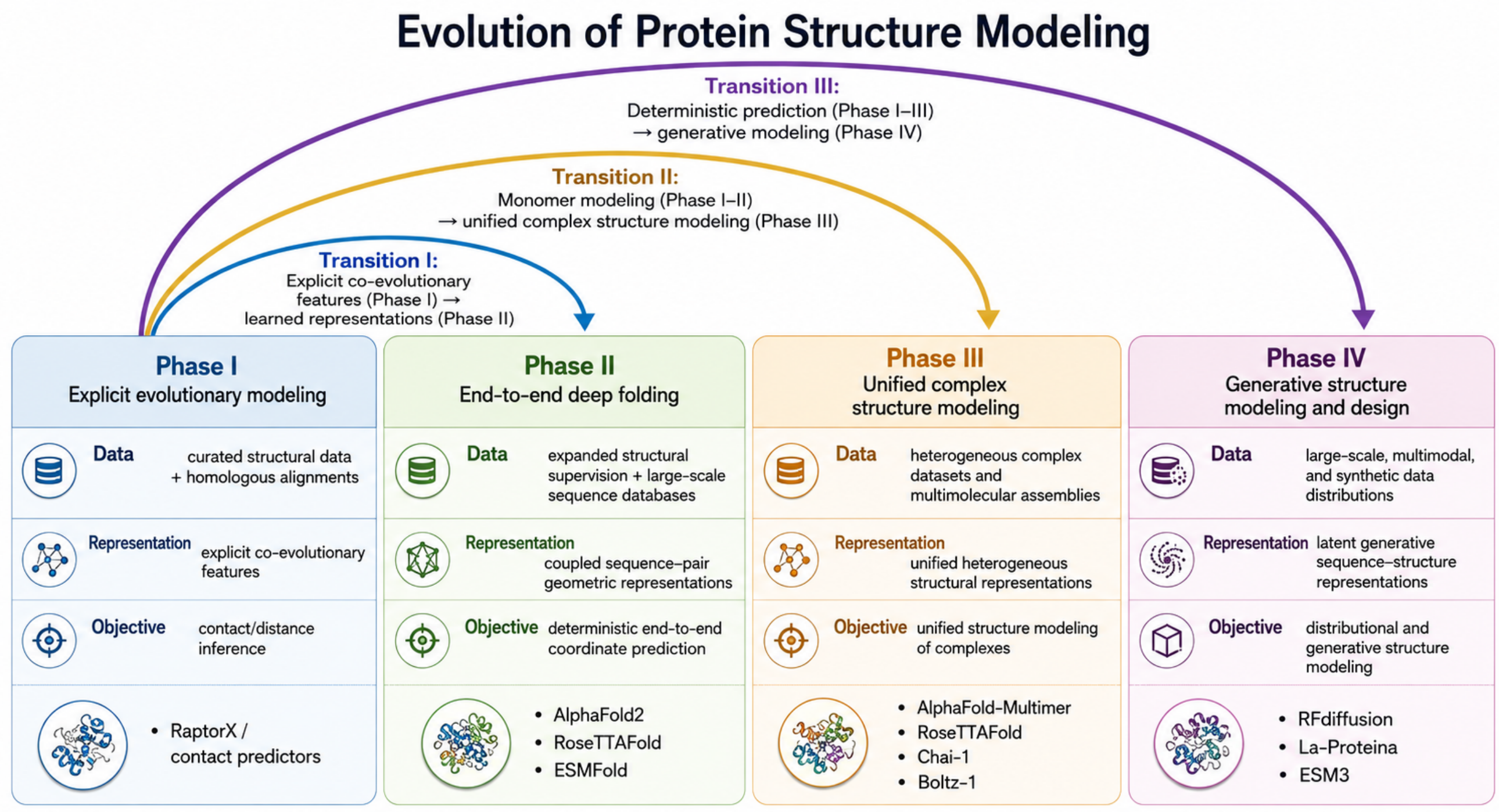

**Figure 1:** Four methodological phases in the evolution of deep learning for protein structure prediction.

## 2. Explicit Evolutionary Modeling

The earliest methodological phase of protein structure prediction with machine learning is characterized by the explicit modeling of evolutionary couplings derived from multiple sequence alignments (MSAs). As illustrated in Figure 2, this paradigm follows a structured and modular pipeline consisting of (A) MSA construction, (B) co-evolutionary feature extraction, (C) contact or distance prediction, and (D) downstream structure reconstruction . The underlying data regime is dominated by curated structural datasets such as the Protein Data Bank (PDB) [17], together with homologous sequence collections obtained through large-scale alignment tools, including ClustalW [45], HMMER [26, 20], HHblits [62], and MMseqs2 [74].

At the core of this phase lies a key assumption: evolutionary constraints encoded in homologous sequences reflect three-dimensional structural relationships. According to the thermodynamic hypothesis of protein folding [6] and subsequent theoretical developments [18, 12], amino acid sequences determine native structures, and residues that are spatially proximal tend to co-evolve to maintain structural stability and function. Statistical approaches such as direct coupling analysis (DCA) quantify these dependencies by estimating residue–residue coupling strengths from MSAs [53, 54, 31]. These coupling signals provide an indirect but powerful proxy for structural proximity and serve as the primary source of structural information in this paradigm.

### 2.1. From Statistical Couplings to Deep Learning-Based Contact Prediction

Building on these statistical foundations, deep learning models introduced more expressive approaches to contact prediction. Convolutional neural networks and residual architectures trained on MSA-derived features were used to predict residue–residue contact maps or distance distributions [82]. In these models, co-evolutionary features—such as covariance matrices, coupling scores, and position-specific profiles—are explicitly computed and used as input representations.

Representative systems such as RaptorX and related frameworks substantially improved long-range contact prediction compared with purely statistical methods [82, 59]. However, despite these improvements, the overall methodological logic remained largely unchanged. Evolutionary signals were still engineered first, predictive constraints were generated second, and full three-dimensional structures were reconstructed only in a downstream stage. In this sense, deep learning enhances predictive accuracy without fundamentally altering the pipeline structure.

### 2.2. Modular Pipeline and Structural Decoupling

As shown in Figure 2, this phase exhibits a distinctly modular architecture. Contact or distance prediction is performed independently of coordinate reconstruction, and predicted constraints are subsequently passed to external folding or reconstruction procedures based on fragment assembly, distance geometry, or knowledge-based potentials [70, 72]. Although some implementations more tightly couple constraint prediction and reconstruction, the dominant early pipeline remained fragmented: gradients did not propagate from final structural outputs through all preceding stages, and individual components were optimized separately.

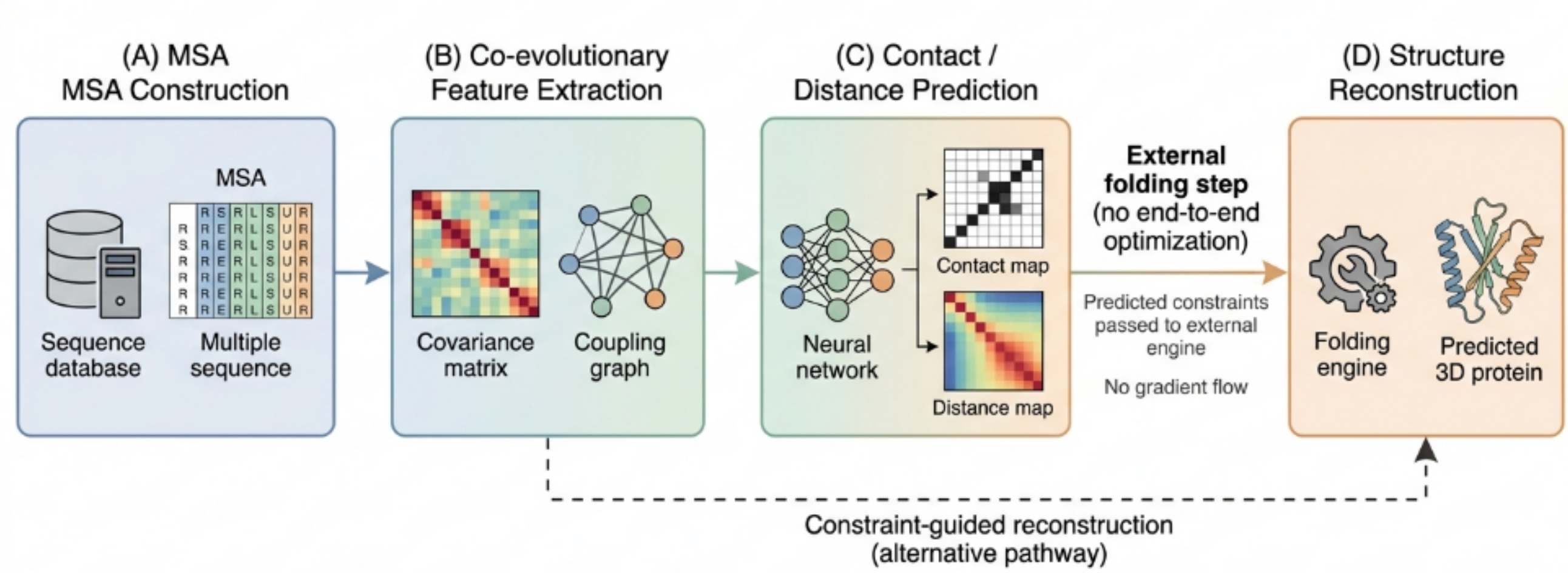

**Figure 2:** Co-evolutionary modeling pipeline in early protein structure prediction. The schematic illustrates (A) MSA construction from sequence databases, (B) extraction of co-evolutionary features (e.g., covariance or coupling matrices), (C) deep learning–based contact or distance prediction, and (D) downstream structure reconstruction using external folding engines.

This design introduces several limitations. Errors in MSA construction, coupling estimation, or contact prediction can propagate into downstream folding without end-to-end correction. Moreover, global geometric consistency is not directly optimized at the contact-prediction stage, so locally plausible constraints may still be incompatible with a physically coherent three-dimensional structure. Consequently, gains in intermediate contact or distance accuracy do not necessarily yield proportional improvements in final structural quality [44, 5].

## 2.3. Methodological Characteristics and System-Level Limitations

These characteristics can be summarized from a system-level perspective, as shown in Table 1. The phase combines curated structural supervision with MSA-dependent homologous sequence information, relies on explicit co-evolutionary features as its principal representation, and formulates prediction in terms of contacts or distances rather than direct coordinate generation.

From a representation standpoint, the strengths and weaknesses of this paradigm are tightly coupled. MSAs [45] encode evolutionary variation through covariance statistics and coupling scores [53, 54, 31], which can provide highly informative signals for structurally conserved protein families. However, predictive performance depends strongly on the depth, diversity, and alignment quality of available homologs [59]; when coverage is limited, co-evolutionary signals become sparse or noisy [82].

This limitation is further reinforced by the data regime. Although sequence databases have expanded rapidly, usable target-specific information remains constrained by the ability to identify and align informative homologs using tools such as HMMER [26, 20], HHblits [62], and MMseqs2 [74]. Proteins with shallow or biased evolutionary coverage therefore remain challenging even when structural supervision is available from repositories such as the PDB [17].

The learning objective introduces an additional limitation. Contact maps and distance matrices are intermediate structural representations rather than final atomic coordinates, and the inverse mapping from a set of predicted constraints to a unique, physically valid three-dimensional structure is generally underdetermined. Reconstruction quality therefore depends on the downstream optimization procedure as well as on the accuracy and consistency of the predicted restraints [46, 78].

## 2.4. Transition Toward End-to-End Modeling

Despite these limitations, explicit evolutionary modeling represents a critical foundation for subsequent developments. It established the importance of pairwise residue relationships, demonstrated that sequence variation can be systematically exploited for structural inference, and provided the conceptual basis for learning structure from sequence-derived constraints.

More importantly, the limitations of this phase—particularly its dependence on explicit feature engineering and its modular separation between prediction and reconstruction—directly motivate the transition to end-to-end deep folding models. Later approaches seek to integrate representation learning, geometric reasoning, and coordinate prediction within a unified optimization framework, thereby addressing the fragmentation and inefficiencies inherent in early pipelines [66, 40].

**Table 1**
Methodological characteristics and limitations of explicit evolutionary modeling in early protein structure prediction.

| Aspect | Description | Limitation |
|---|---|---|
| Data regime | Curated structures + homologous sequences (MSA-dependent) | Limited by MSA depth and coverage |
| Representation | Explicit co-evolutionary features (covariance, coupling) | Requires high-quality alignments |
| Objective | Contact/distance prediction | Indirect relation to final 3D structure |
| Pipeline | Modular (prediction + folding) | Error propagation, no end-to-end optimization |

## 3. From Explicit Evolutionary Features to Learned Sequence Representations

A central methodological transition in protein structure prediction is the shift from explicit evolutionary feature engineering toward learned sequence representations (Figure 1). Rather than relying exclusively on query-specific co-evolutionary statistics derived from MSAs, some modern predictors learn latent representations from large-scale sequence corpora through self-supervised pretraining [32, 22, 47]

### 3.1. MSA-Based Pipelines and Their Underlying Assumptions

In earlier MSA-based pipelines, structural information is obtained through a multi-stage process (Figure 3). Starting from a query sequence, homologous sequences are retrieved from large databases to construct an MSA, which serves as the basis for extracting co-evolutionary features such as covariance matrices and coupling scores. These features are then used as inputs to neural networks for predicting residue–residue contacts or distance distributions, which are subsequently converted into three-dimensional structures through external folding engines or constraint-based reconstruction.

This pipeline reflects the core assumption of explicit evolutionary modeling: structural constraints can be inferred from correlated mutations across homologous sequences. When sufficiently deep and diverse MSAs are available, such co-evolutionary signals provide strong indicators of spatial proximity and functional constraint [53, 54, 31]. Their reliability nevertheless varies with MSA depth, sequence diversity, and alignment quality. Query-specific MSA construction also introduces database-search and preprocessing costs, particularly at proteome scale [74, 56].

### 3.2. Emergence of Learned Sequence Representations

In explicitly MSA-free PLM-based predictors such as ESMFold and OmegaFold, the query sequence is encoded directly by a pretrained protein language model rather than being accompanied by a query-specific MSA. Self-supervised pretraining on large sequence corpora produces contextual embeddings that can encode information relevant to structure and function [47, 87]. The precise structural content and transferability of these embeddings remain model- and task-dependent.

These learned embeddings are then passed to downstream folding or coordinate-prediction modules. In MSA-free implementations, removing query-specific homology search can simplify inference and improve throughput, although the magnitude of the speed and accuracy trade-off varies across models, target classes, and evaluation settings [47, 87]. This paradigm also shifts much of the computational burden from online alignment construction to offline large-scale pretraining.

### 3.3. Pipeline-Level Comparison Between MSA and PLM Paradigms

The contrast between these two paradigms is illustrated in Figure 3. In MSA-based pipelines (Figure 3), structural information is derived from explicitly computed evolutionary features, and the workflow is multi-stage and alignment-dependent. In PLM-based pipelines (Figure 3), structural information is encoded in learned sequence representations, and prediction becomes more integrated and less dependent on external preprocessing. This transition reflects a broader methodological shift: from explicit feature engineering to representation learning. In the former, structural signals are manually derived from sequence variation; in the latter, they are implicitly captured within learned embedding spaces. As a result, the locus of structural information moves from external preprocessing steps to the internal representations of neural networks.

### 3.4. Methodological Differences and Scaling Behavior

From a methodological perspective, the differences between these paradigms can be summarized along several dimensions, as shown in Table 2. MSA-based methods obtain target-specific evolutionary information from homologous sequences, whereas PLM-based methods encode statistical regularities acquired during pretraining. MSA-dependent performance is therefore strongly influenced by homolog availability, while the capabilities of PLM-based systems depend on the architecture, pretraining data, model scale, and downstream folding module. These are tendencies rather

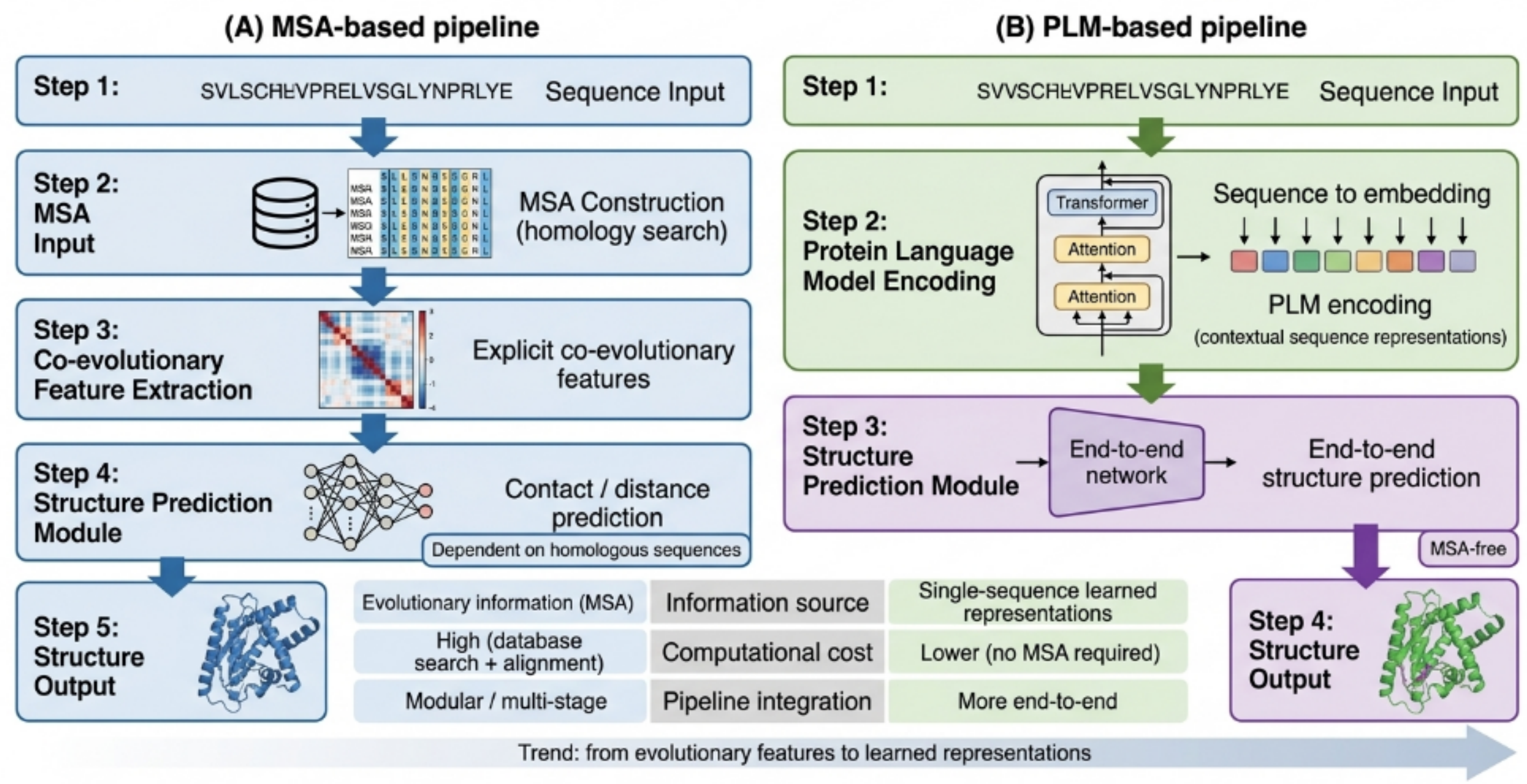

**Figure 3:** Comparison between MSA-based and PLM-based structure prediction pipelines. The schematic contrast (A) traditional pipelines that rely on MSA construction and explicit co-evolutionary feature extraction with (B) PLM-based pipelines that directly encode sequence information into contextual embeddings. Key differences include information source, computational cost, and integration with downstream structure modules.

**Table 2**
Comparison between explicit evolutionary modeling and learned sequence representations

| Dimension | Explicit evolutionary modeling | Learned sequence representations |
|---|---|---|
| Data regime | MSA-dependent homologous sequences | Large-scale unlabeled sequence corpora |
| Representation | Explicit co-evolutionary features (covariance, coupling) | Implicit contextual embeddings (PLMs) |
| Information | Directly computed evolutionary couplings | Learned statistical patterns from sequences |
| Scalability | Limited by MSA depth and coverage | Can improve with pretraining data and model scale, depending on architecture |
| Generalization | Strong when informative homologous coverage is available | Some MSA-free PLM models can be advantageous for orphan or low-homology targets |

than universal properties of every model in either category [15, 24, 47].

These differences lead to distinct but overlapping generalization profiles. MSA-based approaches remain highly effective when informative homologous coverage is available. Some MSA-free PLM predictors, notably ESMFold and OmegaFold, can be useful for orphan or low-homology targets because they do not require a target-specific alignment, but this advantage should not be generalized to all PLM architectures or all difficult proteins [47, 87]. The two information sources are therefore better viewed as complementary than mutually exclusive.

### 3.5. Limitations and Hybrid Strategies

PLM-based approaches introduce their own challenges. Learned embeddings are distributed representations whose biological and geometric content is often less directly interpretable than explicit coupling matrices. In addition, the apparent reduction in inference-time preprocessing can obscure the substantial computational and data requirements of PLM pretraining [21, 16].

Whether learned sequence representations consistently recover the fine-grained residue-pair constraints supplied by deep MSAs remains model- and target-dependent. Hybrid systems that combine PLM embeddings with MSA-derived or pairwise geometric information may therefore be advantageous in some settings [15].

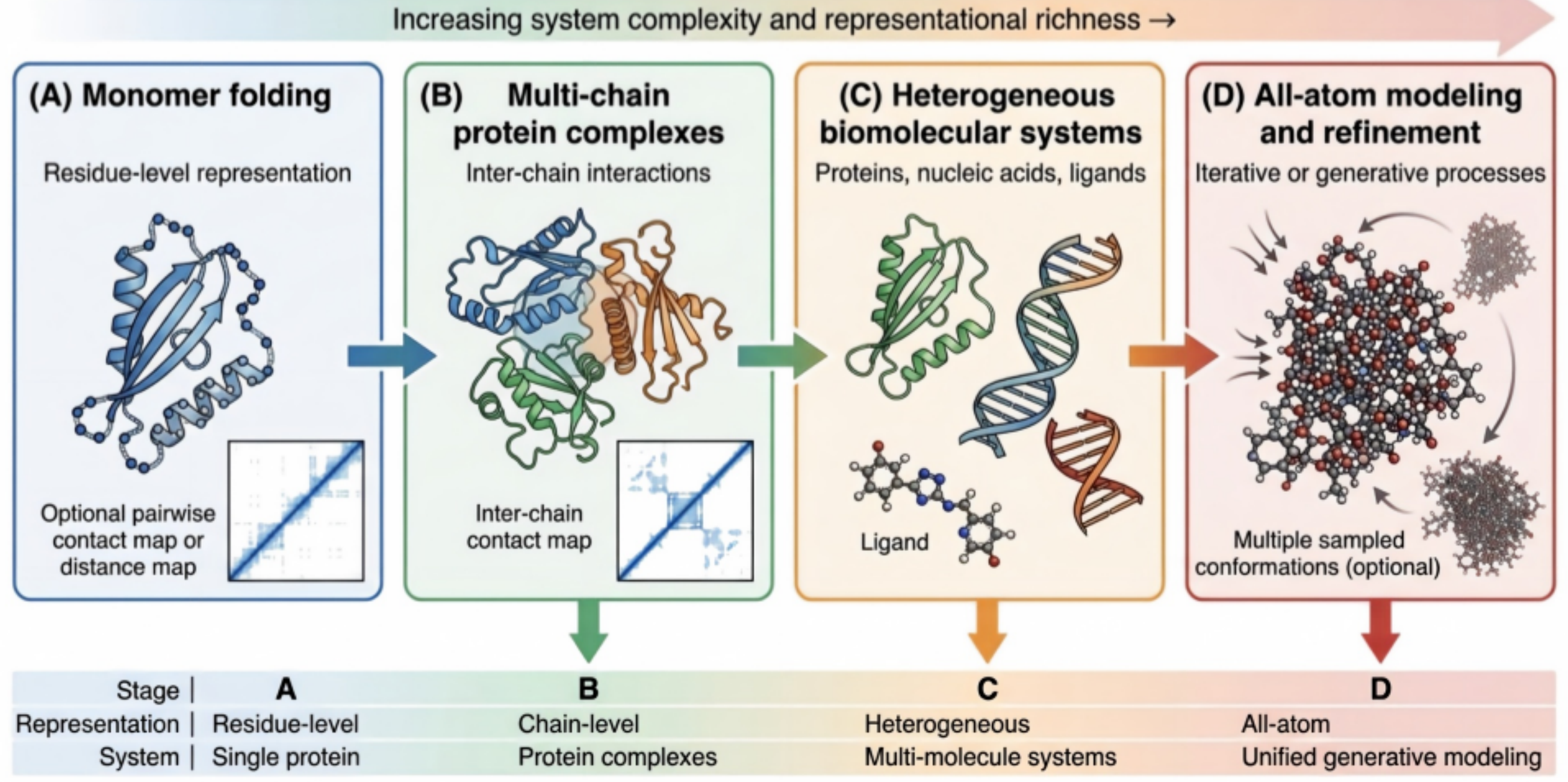

**Figure 4:** Evolution from monomer folding to relatively unified heterogeneous modeling. The schematic illustrates (A) monomer folding with residue-level representation, (B) multi-chain protein-complex modeling, (C) heterogeneous biomolecular systems including nucleic acids and ligands, and (D) mixed token- and atom-level modeling with iterative or diffusion-based refinement.

### 3.6. Transition Toward End-to-End and Unified Modeling

Importantly, this transition should not be interpreted as a simple replacement of MSA-based methods. Rather, it represents a broader shift in how structural information is conceptualized and learned. Evolutionary signals are not discarded but transformed—from explicitly computed statistics to latent representations embedded within model parameters.

More broadly, the move toward learned sequence representations lays the foundation for subsequent methodological developments. By reducing dependence on handcrafted features and enabling more integrated architectures, it facilitates the emergence of end-to-end folding models and, later, unified frameworks for modeling heterogeneous molecular systems. In this sense, this transition serves as a critical bridge between early co-evolution-based approaches and modern deep learning architectures.

## 4. From Monomer Folding to Unified Modeling of Heterogeneous Molecular Assemblies

A major development in contemporary protein structure prediction is the expansion of modeling scope from single-chain folding to the joint treatment of increasingly heterogeneous molecular systems [23, 8]. This transition involves more than an increase in target complexity: it changes the entities represented by the model [42], the geometric and chemical relationships that must be learned, and the criteria by which predictions are evaluated [2]. As illustrated in Figure 4, the evolution of protein structure prediction can be understood as a progressive expansion of both system complexity and representational richness, moving from isolated protein folding toward relatively unified modeling of heterogeneous molecular interactions.

### 4.1. Monomer Folding as the Foundational Paradigm

In the monomer-folding setting, the predictive task can be formulated as mapping a single amino acid sequence to a dominant three-dimensional structure. Such predictions are often assessed using global and local structural-similarity metrics, including TM-score, RMSD, and lDDT; these metrics are not exclusive to monomer prediction, although complex and ligand-containing systems require additional interface- and chemistry-aware measures [92, 52, 10]. This formulation underlies landmark end-to-end models such as AlphaFold2 [40].

Monomer folding nevertheless captures only a subset of biologically relevant structural phenomena. Protein function is frequently determined by binding, multimerization, nucleic-acid recognition, ligand association, and higher-order assembly, motivating models that represent intermolecular interfaces and chemically distinct entities [4, 64].

### 4.2. From Multi-Chain Complexes to Heterogeneous Systems

The first step beyond monomer folding is the extension to multi-chain protein complexes, as exemplified by AlphaFold-Multimer [23]. In this setting, the model must reason about intra-chain folding and inter-chain geometry, including interface arrangement, chain identity, and stoichiometry. Accurate individual-chain structures do not guarantee a correct assembly, so complex prediction requires interface-aware representations, training objectives, and evaluation criteria [93, 13].

The subsequent expansion proceeds through distinct steps. RoseTTAFoldNA extends the three-track framework specifically to protein–DNA and protein–RNA complexes [8], whereas RoseTTAFold All-Atom broadens the modeling space to generalized biomolecular systems that can include ligands and other non-polymer entities [42]. AlphaFold3 further supports joint structure inference for proteins, nucleic acids, ligands, ions, and modified residues within a common predictive system [2]. These models should therefore not be treated as having identical molecular scope.

## 4.3. Toward Relatively Unified Modeling Frameworks

As these capabilities converge, the field moves toward what may be described as relatively unified modeling frameworks. The qualifier "relatively" is important: current systems integrate several previously separate tasks within shared architectures [42, 2], but their accuracy, training coverage, and chemical treatment remain uneven across molecular classes and interaction types [36, 57].

This transition entails a corresponding change in representation. Earlier models emphasized residue-level sequence, pair, and frame representations, whereas heterogeneous systems require mixed token- and atom-level descriptions that encode chemical identity, connectivity, and cross-entity geometry. The predictive objective correspondingly broadens from isolated-chain folding to joint structure inference across interacting molecular components [42, 2].

## 4.4. AlphaFold3 and the Redefinition of the Prediction Problem

Within this phase, AlphaFold3 represents a key methodological turning point [2]. Its significance lies not only in predictive performance, but also in reformulating structure inference around heterogeneous biomolecular interactions.

AlphaFold3 expands the modeled entity space by supporting proteins, nucleic acids, ligands, ions, and modified residues within a shared predictive framework. This is methodologically important because many biological interactions depend on precise geometry between chemically distinct components rather than on protein folds considered in isolation [2].

Its representation and output formulation also move toward joint atom-level modeling of heterogeneous systems, enabling the network to reason about local geometry, chemical identity, and interaction-specific contacts within a common structure-prediction pipeline [2]. The term "all-atom" should not be interpreted as implying perfect stereochemical validity in every prediction.

AlphaFold3 replaces the AlphaFold2 structure module with a diffusion module that denoises atom coordinates conditioned on processed input representations. Repeated random seeds can yield multiple candidate predictions, but the model remains prediction-oriented because the identities of the molecular entities are specified and the primary objective is to infer their joint structure. The presence of a diffusion architecture therefore does not, by itself, place a model in the design-oriented generative phase [2].

AlphaFold3 also increases the importance of evaluating local chemical validity. The diffusion formulation accommodates diverse chemical components and reduces reliance on molecule-specific output modules, yet residual stereochemical violations and clashes can still occur. Structural accuracy should therefore be assessed using both global or interface-level similarity and local chemistry-aware validation [2, 14].

## 4.5. Methodological Implications and Remaining Challenges

The transition from monomer folding to heterogeneous modeling introduces systematic changes across multiple aspects of the problem. These changes extend beyond representation and encompass the definition of prediction targets, the nature of available data, and the criteria used for evaluation.

To clarify these differences, a structured comparison between monomer folding and heterogeneous molecular system modeling is provided in Table 3.

These changes introduce several challenges. High-quality training and benchmark data are unevenly distributed across complex types; representations must balance chemical detail with computational tractability; and evaluation must distinguish global assembly accuracy from interface fidelity, ligand pose quality, and local stereochemical validity [51, 19, 89].

Furthermore, broader modeling scope does not eliminate the value of specialized inductive biases. Antibody modeling, particularly in highly variable CDR loops, remains a prominent example in which domain-specific systems such as IgFold can offer practical advantages [1, 65].

## 4.6. Transition Toward Generative Modeling

The progression summarized in Figure 4 also highlights a broader methodological trajectory. As models become capable of representing heterogeneous molecular entities within shared all-atom spaces and performing iterative refinement, the conceptual boundary between prediction and generation begins to diminish.

In this sense, the expansion toward heterogeneous modeling serves as a bridge to subsequent developments in generative structural modeling. By redefining the prediction target from isolated protein folds to interaction-rich molecular systems, this phase lays the foundation for approaches that treat structure not as a single deterministic outcome, but as part of a broader distribution of plausible configurations.

**Table 3**
Comparison between monomer folding and heterogeneous molecular system modeling. The table contrasts the two phases in terms of system scope, representation, objective, data regime, and evaluation, highlighting the movement from single-chain residue-level prediction toward interaction-aware modeling of multi-molecular assemblies.

| Dimension | Monomer folding (Phase II) | Heterogeneous modeling (Phase III) |
|---|---|---|
| System scope | Single protein chain | Multi-chain and multimolecular assemblies |
| Representation | Residue-level sequence–pair and frame representations | Mixed token and atom-level heterogeneous representations |
| Objective | Structure inference for a specified single-chain protein | Joint structure inference for specified multimolecular assemblies |
| Data regime | Relatively richer monomer-focused structural data | More heterogeneous and unevenly sampled complex data |
| Evaluation | Global and local structural-similarity metrics | Global, interface-specific, and chemistry-aware metrics |

## 5. From Prediction-Oriented Structure Inference to Generative Protein Design

As illustrated in the revised Figure 4, the key distinction is not whether a model produces one or multiple samples, nor whether it uses a diffusion module. It is whether the primary objective is inference for specified molecular entities or generation of novel design candidates under constraints.

### 5.1. Prediction-Oriented Structure Inference

Prediction-oriented structure models estimate conformations for specified molecular inputs. AlphaFold2 largely behaves as a point-estimation system that returns a ranked structural hypothesis with confidence metrics [40]. AlphaFold3, by contrast, uses diffusion-based coordinate denoising and can generate different candidate predictions across seeds, while still remaining prediction-oriented because the sequence and molecular composition are given [2].

Confidence measures such as pLDDT [77] and predicted aligned error [2] quantify expected reliability of predicted regions or relative placements, but they should not be equated with a complete conformational ensemble [3, 84]. Likewise, variability across random seeds may reflect model sampling behavior [2] without necessarily reproducing the biologically populated distribution of states [84].

Prediction-oriented inference is highly effective when the scientific question concerns the likely structure of a specified target under conditions represented by the training data. Its limitations become more apparent when the target occupies multiple biologically relevant states or when context-dependent dynamics are central to function.

### 5.2. Structural Heterogeneity and Distributional Prediction

Many proteins and complexes occupy conformational ensembles shaped by thermodynamic fluctuations, molecular partners, and environmental conditions. A single high-confidence structural hypothesis may therefore be insufficient for intrinsically disordered regions, fold-switching proteins, flexible loops, or ligand-induced rearrangements [11, 58, 76].

This problem motivates a distinct line of work on distributional structure prediction, in which a model samples alternative conformations for a specified sequence. Such approaches should be distinguished from de novo protein design: both can use generative machinery, but one aims to approximate a target's conformational distribution, whereas the other aims to create new molecular solutions. EigenFold is one representative example of sequence-conditioned distributional structure prediction [39].

Accordingly, neither point prediction nor repeated stochastic sampling should automatically be interpreted as a faithful physical ensemble. Validation against experimental ensemble data, molecular simulations, or state-specific benchmarks remains necessary [37, 68].

### 5.3. Generative Modeling for Protein Design

Generative protein-design models are defined primarily by their ability to create novel molecular candidates rather than merely by their ability to sample multiple structures. Depending on the task, they may generate protein backbones, amino acid sequences, side-chain arrangements, or joint sequence–structure representations under design constraints.

Diffusion models [30] have become an influential design paradigm because conditional denoising can be guided by geometric motifs, target surfaces, symmetries, or other structural constraints [83]. Flow-matching [48] and autoregressive approaches [81] provide alternative routes to the same broad objective: learning a generative process over molecular design space.

RFdiffusion generates novel protein backbones conditioned on structural or functional constraints and has been applied to motif scaffolding, symmetric assemblies, and binder design [83]. Its primary contribution is therefore controllable de novo design rather than conformational sampling of a fixed natural sequence.

La-Proteina extends this direction through partially latent flow matching for atomistic protein generation and joint treatment of sequence–structure information [27]. RFdiffusion and La-Proteina differ in representation and generation strategy, but both are better characterized as design-oriented generative systems than as generic uncertainty models.

**Table 4**
Comparison between prediction-oriented structure inference and generative protein design. The table distinguishes the paradigms by conditioning, primary objective, output, and modeling mechanism; it deliberately avoids equating prediction with a single output or generation with multiple samples.

| Dimension | Prediction-oriented structure inference | Generative protein design |
|---|---|---|
| Conditioning | Specified sequence and/or molecular composition | Motifs, target geometry, interfaces, function, and optional sequence constraints |
| Primary objective | Infer structures compatible with specified molecules | Create novel structures or sequence–structure solutions |
| Output | One or more candidate structures for the same specified target | Diverse de novo backbones, sequences, or joint designs |
| Inference/generation mechanism | Deterministic regression or diffusion-based sampling | Diffusion, flow matching, autoregressive, or other generative processes |
| Representative models | AlphaFold2, AlphaFold3 | RFdiffusion, La-Proteina |

### 5.4. Conceptual Differences Between Prediction and Generation

The distinction between prediction-oriented inference and generative design therefore concerns conditioning and objective. Prediction starts from specified molecular identities and asks what structure or interaction geometry is plausible. Generative design starts from a desired motif, target interface, geometry, or functional objective and searches for novel molecular realizations. Either paradigm may produce multiple candidates, and either may use diffusion.

This distinction has direct implications for evaluation. Predictive models are assessed primarily by correspondence to experimentally determined structures and by confidence calibration, whereas design models must additionally be evaluated for novelty, validity, diversity, designability, and experimental function [43, 49, 90].

To clarify these differences, a structured comparison is provided in Table 4.

### 5.5. Complementarity and Emerging Unified Perspective

Prediction-oriented inference and generative design are complementary. Predictive models can evaluate whether a designed sequence is compatible with a target fold [40, 83] or interaction geometry [2], while generative models can propose candidates that satisfy specified constraints [83]. In practical design loops, prediction, generation, ranking, and experimental validation may therefore be iterated rather than treated as isolated stages [61, 71].

This complementarity contributes to a broader convergence between structural inference and protein engineering. Nevertheless, prediction of a specified structure and generation of a novel design remain different tasks, with different notions of success and different validation requirements [83, 27].

### 5.6. Challenges and Future Directions

Generative protein design introduces several challenges. Generated candidates must satisfy stereochemical, energetic, and folding constraints; computational metrics may not predict expression, stability, binding, or biological function; and diversity must be distinguished from invalid or redundant sampling. Evaluation therefore requires a combination of structural validation, sequence analysis, physical modeling, and experimental testing. [90, 9] Computational cost and sampling efficiency also remain important, especially for iterative diffusion or flow-based systems [34, 91].

Despite these challenges, generative modeling represents a fundamental expansion of computational structural biology. It extends the field from inference about specified molecules toward the controlled creation and exploration of new molecular solutions. This design-oriented phase complements, rather than supersedes, predictive structure models.

## 6. Synthesis, Challenges, and Future Directions

### 6.1. Synthesis of methodological evolution

Protein structure prediction has undergone a profound methodological transformation over the past decades, evolving from physics-based and knowledge-based modeling to deep learning–driven inference and, more recently, to increasingly integrated and generative frameworks. As organized in this review, this development is best understood not simply as a sequence of progressively stronger models, but as a broader reconfiguration of how structural information is represented, learned, and utilized. The four methodological phases and three cross-cutting transitions discussed throughout the review provide a structured way to interpret this evolution beyond a purely chronological narrative.

At a fundamental level, the history of protein structure prediction reflects a continuing redefinition of what counts as the relevant source of structural information. Early approaches relied on explicit physical principles, statistical potentials, and optimization over energy landscapes [18, 6, 12, 72]. The subsequent rise of co-evolutionary analysis introduced a new form of sequence-derived structural signal, enabling residue-level constraints to be inferred from variation across homologous sequences [53, 54, 31]. In this regime, structural reasoning was still mediated by explicitly constructed features and modular pipelines, but the field had already shifted toward data-informed inference.

The integration of deep learning marked a decisive methodological turning point. Models such as AlphaFold2 and RoseTTAFold [40, 7] demonstrated that structure prediction could be reformulated as an end-to-end learning problem in which representation learning, geometric reasoning, and coordinate prediction are tightly coupled. In parallel, protein language models introduced a complementary route in which structural information was no longer extracted primarily through explicit evolutionary features, but learned implicitly from large-scale sequence corpora [47]. Together, these developments changed both the data regime and the representational logic of the field, shifting the emphasis from handcrafted evolutionary descriptors to learned internal representations.

A further transformation occurred when the scope of prediction expanded beyond isolated monomer folding. Models such as AlphaFold-Multimer [23], RoseTTAFoldNA [8], RoseTTAFold All-Atom [42], and AlphaFold3 [2] moved the field toward increasingly integrated modeling of complexes and heterogeneous molecular systems. In this setting, the target of prediction is no longer just a single protein fold, but the structural organization of multiple interacting entities, often including nucleic acids, ligands, and ions. This shift is particularly important because it aligns structure prediction more closely with biological reality, where function is frequently determined by interfaces, assemblies, and molecular context rather than by isolated folds alone.

More recently, design-oriented generative modeling has introduced another conceptual change. RFdiffusion and La-Proteina use diffusion or flow-matching mechanisms to generate novel backbones or sequence–structure solutions under design constraints [83, 27]. This direction extends structural modeling from inference toward generation, optimization, and molecular engineering.

Taken together, these developments suggest that the recent history of protein structure prediction is not merely a story of accuracy improvement. It is also a story of methodological expansion: from explicit evolutionary modeling to learned representations, from single-structure prediction to uncertainty-aware generation, and from isolated proteins to heterogeneous biomolecular systems. The phase-and-transition framework adopted in this review is intended to make these deeper shifts visible and to clarify how they shape the differing capabilities and practical roles of contemporary models.

## 6.2. Current challenges and unresolved issues

Despite these advances, several important challenges remain. The first concerns the availability and distribution of data. Monomer-focused structure prediction benefits from relatively rich structural and sequence resources, whereas high-quality examples of heterogeneous complexes, alternative conformations, and chemically diverse interactions are more unevenly distributed. This imbalance can bias training and evaluation toward well-represented subsets of biomolecular space [41, 67].

A second challenge concerns representation and interpretability. Explicit co-evolutionary features can be related to residue-pair constraints, whereas latent representations in PLMs, all-atom predictors, and generative models are harder to map onto biological mechanisms. Improved mechanistic interpretation would support model validation, failure analysis, and appropriate scientific use [33, 69, 79].

A third unresolved issue concerns evaluation. Metrics such as TM-score, RMSD, and lDDT remain useful across monomer and complex tasks, but they are insufficient on their own as modeling scope broadens. Complex prediction requires interface-aware assessment [55]; ligand-containing systems require pose and chemistry validation [14]; distributional prediction requires ensemble-aware benchmarks [38, 73]; and generative design requires measures of validity, novelty, diversity, designability, and experimental function [43, 49, 90, 9].

A fourth challenge concerns computational cost and accessibility. MSA-free inference reduces query-time database search in some systems, but large PLMs incur substantial pretraining and memory costs. Heterogeneous all-atom prediction and iterative generative sampling can also increase inference expense. Broader deployment therefore requires transparent reporting of hardware, preprocessing, sampling count, and runtime [16, 94].

Finally, an important conceptual challenge concerns the relationship between generality and specialization. Broader architectures do not necessarily dominate every specialized setting. Antibody modeling, especially for variable CDR regions, illustrates how domain-focused training and inductive biases may remain valuable [60, 65, 86]. Determining when a general model is sufficient and when specialized modeling is justified will remain an important practical question.

## 6.3. Future directions and emerging opportunities

Several directions are likely to shape the next phase of development. One is tighter integration of prediction and design in iterative computational–experimental workflows, in which generative models propose candidates and predictive or experimental modules evaluate structural and functional constraints [28, 83].

Another important direction is the incorporation of dynamics, ensembles, and context dependence. Most current predictors emphasize static structural hypotheses, whereas many proteins function through conformational transitions or partner-dependent states. Progress will require ensemble-aware training data, experimentally grounded benchmarks, and integration with statistical mechanics or molecular simulation [38, 73].

A third opportunity lies in multimodal integration. Structural predictions can potentially be constrained by cryo-EM

density, cross-linking, mutational scans, biochemical assays, ligand-binding measurements, and other experimental evidence [4, 64, 88, 25].

A fourth likely direction is the development of hybrid frameworks that combine prediction distributional inference, and generative design without conflating their objectives. Predictive models may evaluate target compatibility [40, 2], distributional models may explore alternative states [39, 37], and design models may propose novel molecular solutions [83, 35].

Finally, future models will need greater emphasis on efficiency, robustness, reproducibility, and accessibility. Useful reporting standards should include training-data provenance, inference settings, sampling counts, uncertainty calibration, and hardware requirements [29, 80].

### 6.4. Closing perspective

Protein structure prediction is no longer a narrowly defined problem of mapping one sequence to one static fold. It has become a broader modeling discipline that spans learned representation, interaction-aware structural reasoning, and increasingly generative approaches to design and exploration. The methodological evolution reviewed here shows that recent progress is best understood not only in terms of benchmark performance, but also in terms of changing assumptions about data, representation, objective, and scope.

Seen from this perspective, the field is moving from a protein-centric prediction problem toward a more general framework for modeling biomolecular organization. At the same time, its future progress will depend on how effectively it addresses the challenges of data imbalance, interpretability, evaluation, and computational accessibility. The boundary between prediction and design is likely to become increasingly blurred, and the most influential future models may be those that integrate deterministic inference, heterogeneous system modeling, and generative sampling within coherent and practically usable frameworks. In that sense, the recent history of protein structure prediction is not only a record of rapid technical progress, but also a guide to the conceptual directions that are likely to shape its next phase.

## Acknowledgements

The authors are grateful to the support of the Guangdong Key Disciplines Project (2024ZDJS137).

## Declaration of competing interest

The authors declare that they have no known competing financial interests or personal relationships that could have appeared to influence the work reported in this paper.

## CRediT authorship contribution statement

**Wengan He:** Conceptualization, Methodology, Investigation, Data curation, Visualization, Writing – original draft . **Yongsheng Luo:** Conceptualization, Methodology, Investigation, Writing – original draft . **Lihong Jiang:** Validation, Writing – review & editing . **Wenhui Xu:** Validation, Writing – review & editing . **Yu Li:** Conceptualization, Supervision, Writing – review & editing .

## Data availability statement

No new datasets were generated or analyzed in this review. All information discussed in the article was obtained from publicly available publications, databases, and resources cited in the manuscript.

## Declaration of generative AI and AI-assisted technologies in the manuscript preparation process

During the preparation of this work, the authors used ChatGPT, provided by OpenAI, to assist with language refinement, structural organization, and the drafting of selected manuscript and submission-related text. After using this tool, the authors reviewed and edited the content as needed and take full responsibility for the content of the article.